\documentclass[11pt,a4paper]{article}
\usepackage[hyperref]{naaclhlt2018}
\usepackage{times}
\usepackage{latexsym}

\usepackage{url}

\aclfinalcopy 
\def\aclpaperid{972} 

\usepackage{amsmath}
\usepackage{tabularx}
\usepackage{booktabs}
\usepackage{amssymb}
\usepackage{pifont}
\usepackage{bm}
\usepackage{multirow}
\usepackage{tcolorbox}
\usepackage{enumitem}
\usepackage[framemethod=tikz]{mdframed}
\usepackage{tikz,pgfplots}
\usetikzlibrary{matrix}
\usepgfplotslibrary{groupplots}
\pgfplotsset{compat=1.13}

\newcommand\muc[0] {MUC}
\newcommand\bcubed[0] {$\text{B}^3$}
\newcommand\ceaf[0] {$\text{CEAF}_{\phi_4}$}

\newcommand{\M}[2]{\mathbf{#1}_{\text{#2}}}
\newcommand{\V}[1]{\bm{#1}}
\newcommand{\mscore}[1]{s_\text{m}(#1)}
\newcommand{\ascore}[2]{s_\text{a}(#1, #2)}
\newcommand{\bscore}[2]{s_\text{c}(#1, #2)}
\newcommand{\cscore}[2]{s(#1, #2)}
\newcommand{\ffnn}[2]{\textsc{ffnn}_\text{#1}(#2)}

\definecolor{g-red}{HTML}{DB4437}
\definecolor{g-blue}{HTML}{4285F4}
\definecolor{g-green}{HTML}{0F9D58}
\definecolor{g-yellow}{HTML}{F4B400}
\definecolor{g-orange}{HTML}{FF9800}
\definecolor{g-grey}{HTML}{9E9E9E}
\definecolor{uw}{RGB}{138,43,226}
\definecolor{stanford}{RGB}{255,69,0}
\definecolor{const}{RGB}{68, 110, 182}
\definecolor{head}{RGB}{246, 180, 32}
\definecolor{freq}{RGB}{0, 0, 0}

\newmdenv[innerlinewidth=0.5pt, roundcorner=4pt,linecolor=black,innerleftmargin=6pt,
innerrightmargin=6pt,innertopmargin=6pt,innerbottommargin=6pt]{examplebox}

\title{Higher-order Coreference Resolution with Coarse-to-fine Inference}

\author{Kenton Lee \qquad Luheng He \qquad Luke Zettlemoyer\\
  Paul G. Allen School of Computer Science \& Engineering \\
  University of Washington, Seattle WA \\
  {\tt \{kentonl, luheng, lsz\}@cs.washington.edu}}
\date{}

\begin{document}
\maketitle
\begin{abstract}
We introduce a fully differentiable approximation to higher-order inference for coreference resolution. Our approach uses the antecedent distribution from a span-ranking architecture as an attention mechanism to iteratively refine span representations. This enables the model to softly consider multiple hops in the predicted clusters. To alleviate the computational cost of this iterative process, we introduce a coarse-to-fine approach that incorporates a less accurate but more efficient bilinear factor, enabling more aggressive pruning without hurting accuracy. Compared to the existing state-of-the-art span-ranking approach, our model significantly improves accuracy on the English OntoNotes benchmark, while being far more computationally efficient.
\end{abstract}

\section{Introduction}
Recent coreference resolution systems have heavily relied on first order models~\cite{clark:2016b,e2e-coref}, where only pairs of entity mentions are scored by the model. These models are computationally efficient and scalable to long documents. However, because they make independent decisions about coreference links, they are susceptible to predicting clusters that are locally consistent but globally inconsistent. Figure~\ref{fig:consistency} shows an example from \newcite{wiseman:2016} that illustrates this failure case. The plurality of \textbf{[you]} is underspecified, making it locally compatible with both \textbf{[I]} and \textbf{[all of you]}, while the full cluster would have mixed plurality, resulting in global inconsistency.

We introduce an approximation of higher-order inference that uses the span-ranking architecture from \newcite{e2e-coref} in an iterative manner. At each iteration, the antecedent distribution is used as an attention mechanism to optionally update existing span representations, enabling later coreference decisions to softly condition on earlier coreference decisions. For the example in Figure~\ref{fig:consistency}, this enables the linking of \textbf{[you]} and \textbf{[all of you]} to depend on the linking of \textbf{[I]} and \textbf{[you]}.

To alleviate computational challenges from this higher-order inference, we also propose a coarse-to-fine approach that is learned with a single end-to-end objective. We introduce a less accurate but more efficient coarse factor in the pairwise scoring function. This additional factor enables an extra pruning step during inference that reduces the number of antecedents considered by the more accurate but inefficient fine factor. Intuitively, the model cheaply computes a rough sketch of \emph{likely} antecedents before applying a more expensive scoring function.

Our experiments show that both of the above contributions improve the performance of coreference resolution on the English OntoNotes benchmark. We observe a significant increase in average F1 with a second-order model, but returns quickly diminish with a third-order model. Additionally, our analysis shows that the coarse-to-fine approach makes the model performance relatively insensitive to more aggressive antecedent pruning, compared to the distance-based heuristic pruning from previous work.

\begin{figure}[t!]
\centering
\setlength{\fboxsep}{0.5em}
\fbox{\parbox{0.8\linewidth}{
\textit{Speaker 1}:  Um and \textbf{[I]} think that is what's - Go ahead Linda.\\
\textit{Speaker 2}:  Well and uh thanks goes to \textbf{[you]} and to the media to help us... So our hat is off to \textbf{[all of you]} as well.
}}
\caption{Example of consistency errors to which first-order span-ranking models are susceptible. Span pairs (\textbf{I}, \textbf{you}) and (\textbf{you}, \textbf{all of you}) are locally consistent, but the span triplet (\textbf{I}, \textbf{you}, \textbf{all of you}) is globally inconsistent. Avoiding this error requires modeling higher-order structures.}
\label{fig:consistency}
\end{figure}
\section{Background}
\paragraph{Task definition}
We formulate the coreference resolution task as a set of antecedent assignments $y_i$ for each of span $i$ in the given document, following \newcite{e2e-coref}. The set of possible assignments for each $y_i$ is $\mathcal{Y}(i) = \{\epsilon, 1, \ldots, i - 1\}$, a dummy antecedent $\epsilon$ and all preceding spans. Non-dummy antecedents represent coreference links between $i$ and $y_i$. The dummy antecedent $\epsilon$ represents two possible scenarios: (1) the span is not an entity mention or (2) the span is an entity mention but it is not coreferent with any previous span. These decisions implicitly define a final clustering, which can be recovered by grouping together all spans that are connected by the set of antecedent predictions.  
 
\paragraph{Baseline}
We describe the baseline model~\cite{e2e-coref}, which we will improve to address the modeling and computational limitations discussed previously. The goal is to learn a distribution $P(y_i)$ over antecedents for each span $i$ :
\begin{align}
P(y_i) &= \frac{e^{s(i, y_i)}}{\sum_{y' \in \mathcal{Y}(i)}e^{s(i, y')}}
\end{align}
where $s(i, j)$ is a pairwise score for a coreference link between span $i$ and span $j$. The baseline model includes three factors for this pairwise coreference score: (1) $\mscore{i}$, whether span $i$ is a mention, (2) $\mscore{j}$, whether span $j$ is a mention, and (3) $\ascore{i}{j}$ whether $j$ is an antecedent of $i$:
\begin{align}
    \cscore{i}{j} &=\mscore{i} + \mscore{j} + \ascore{i}{j}
\end{align}
In the special case of the dummy antecedent, the score $s(i, \epsilon)$ is instead fixed to 0. A common component used throughout the model is the vector representations $\V{g}_i$ for each possible span $i$. These are computed via bidirectional LSTMs~\cite{lstm} that learn context-dependent boundary and head representations. The scoring functions $s_\text{m}$ and $s_\text{a}$ take these span representations as input:
\begin{align}
    \mscore{i}&= \V{w}_\text{m}^\top \ffnn{m}{\V{g}_i}\\
\hspace{-10pt}\ascore{i}{j} &= \V{w}_\text{a}^\top \ffnn{a}{[\V{g}_i, \V{g}_j, \V{g}_i \circ \V{g}_j, \phi(i, j)]}\hspace{-10pt}
\end{align}
\noindent
where $\circ$ denotes element-wise multiplication, $\textsc{ffnn}$ denotes a feed-forward neural network, and the antecedent scoring function $\ascore{i}{j}$ includes explicit element-wise similarity of each span $\V{g}_i \circ \V{g}_j$ and a feature vector $\phi(i, j)$ encoding speaker and genre information from the metadata and the distance between the two spans.

The model above is factored to enable a two-stage beam search. A beam of up to $M$ potential mentions is computed (where $M$ is proportional to the document length) based on the spans with the highest mention scores $\mscore{i}$. Pairwise coreference scores are only computed between surviving mentions during both training and inference.

Given supervision of gold coreference clusters, the model is learned by optimizing the marginal log-likelihood of the possibly correct antecedents. This marginalization is required since the best antecedent for each span is a latent variable.

\section{Higher-order Coreference Resolution}
\label{sec:higher_order}
The baseline above is a first-order model, since it only considers pairs of spans. First-order models are susceptible to consistency errors as demonstrated in Figure~\ref{fig:consistency}. Unlike in sentence-level semantics, where higher-order decisions can be implicitly modeled by the LSTMs, modeling these decisions at the document-level requires explicit inference due to the potentially very large surface distance between mentions.

We propose an inference procedure that allows the model to condition on higher-order structures, while being fully differentiable. This inference involves $N$ iterations of refining span representations, denoted as $\V{g}_i^n$ for the representation of span $i$ at iteration $n$. At iteration $n$, $\V{g}_i^n$ is computed with an attention mechanism that averages over previous representations $\V{g}_j^{n-1}$ weighted according to how likely each mention $j$ is to be an antecedent for $i$, as defined below. 

The baseline model is used to initialize the span representation at $\V{g}_i^1$. The refined span representations allow the model to  also iteratively refine the antecedent distributions $P_n(y_i)$:
\begin{align}
P_n(y_i) &= \frac{e^{s(\V{g}_i^n, \V{g}_{y_i}^n)}}{\sum_{y \in \mathcal{Y}(i)}e^{s(\V{g}_i^n, \V{g}_y^n))}}
\end{align}
where $s$ is the coreference scoring function of the baseline architecture. The scoring function uses the same parameters at every iteration, but it is given different span representations.

At each iteration, we first compute the expected antecedent representation $\V{a}_i^n$ of each span $i$ by using the current antecedent distribution $P_{n}(y_i)$ as an attention mechanism:
\begin{align}
\V{a}_i^n &= \sum_{y_i \in \mathcal{Y}(i)}P_{n}(y_i) \cdot \V{g}_{y_i}^n
\end{align}

The current span representation $\V{g}_i^n$ is then updated via interpolation with its expected antecedent representation $\V{a}_i^n$:
\begin{align}
\V{f}_i^n &= \sigma(\M{W}{f}[\V{g}_i^n, \V{a}_i^n]) \\
\V{g}_i^{n+1}&= \V{f}_i^n \circ \V{g}_i^n + (\V{1} - \V{f}_i^n) \circ \V{a}_i^n
\end{align}
The learned gate vector $\V{f}_i^n$ determines for each dimension whether to keep the current span information or to integrate new information from its expected antecedent.
At iteration $n$, $\V{g}_i^n$ is an element-wise weighted average of approximately $n$ span representations (assuming $P_n(y_i)$ is peaked), allowing $P_n(y_i)$ to softly condition on up to $n$ other spans in the predicted cluster.

Span-ranking can be viewed as predicting latent antecedent trees~\cite{fernandes:2012,martschat:2015}, where the predicted antecedent is the parent of a span and each tree is a predicted cluster. By iteratively refining the span representations and antecedent distributions, another way to interpret this model is that the joint distribution $\prod_i P_N(y_i)$ implicitly models every directed path of up to length $N + 1$ in the latent antecedent tree.

\section{Coarse-to-fine Inference}
\label{sec:c2f}
The model described above scales poorly to long documents. Despite heavy pruning of potential mentions, the space of possible antecedents for every surviving span is still too large to fully consider. The bottleneck is in the antecedent score $\ascore{i}{j}$, which requires computing a tensor of size $M \times M \times (3|\V{g}| + |\phi|)$.

This computational challenge is even more problematic with the iterative inference from Section~\ref{sec:higher_order}, which requires recomputing this tensor at every iteration.

\subsection{Heuristic antecedent pruning}
To reduce computation, \newcite{e2e-coref} heuristically consider only the nearest $K$ antecedents of each span, resulting in a smaller input of size $M \times K \times (3|\V{g}| + |\phi|)$.

The main drawback to this solution is that it imposes an a priori limit on the maximum distance of a coreference link. The previous work only considers up to $K = 250$ nearest mentions, whereas coreference links can reach much further in natural language discourse.

\begin{figure}[t!]
\small
\begin{tikzpicture}
\begin{groupplot}[
        group style={
          group name=my plots,
          group size=1 by 1,
          xlabels at=edge bottom,
          xticklabels at=edge bottom,
          vertical sep=0pt
        },
    	width=\linewidth,
	    height=0.7\linewidth,
	    legend cell align=left,
	    legend style={at={(1, 0)},anchor=south east,font=\small},
	    xtick={50, 100, 150, 200, 250},
   		xtick pos=left,
   		xtick align=outside,
	    xmin=25,xmax=275,
		ylabel style={align=center,text width=3cm},
   	 	ymajorgrids=true,
    	xmajorgrids=true,
    	xlabel=Antecedents per span $K$]
\nextgroupplot[
    ymin=67, 
    ymax=73.5, 
    ytick={68, 70, 72},
    ylabel=Dev. Avg. F1 (\%)]
\addplot[
    color=g-blue,
    line width=1.5pt
    ]
    coordinates {(50,72.6)(100,72.7)(250,72.8)};
    \addlegendentry{Coarse-to-fine (various $K$)}
\addplot[
    only marks,
    color=g-blue,
    mark=*,
    mark size=3pt
    ]
    coordinates {
(50,72.6)
    };
    \addlegendentry{Coarse-to-fine (chosen $K$)}
\addplot[
    color=g-red,
    line width=1.5pt
    ]
    coordinates {(50,67.8)(100,71.0)(250,72.5)};
    \addlegendentry{Heuristic (various $K$)}
\addplot[
    only marks,
    color=g-red,
    mark=*,
    mark size=3pt
    ]
    coordinates {
(250,72.5)
    };
    \addlegendentry{Heuristic (chosen $K$)}
\end{groupplot}
\end{tikzpicture}
\caption{Comparison of accuracy on the development set for the two antecedent pruning strategies with various beams sizes $K$. The distance-based heuristic pruning performance drops by almost 5 F1 when reducing $K$ from 250 to 50, while the coarse-to-fine pruning results in an insignificant drop of less than 0.2 F1.}
\label{fig:pruning}
\end{figure}
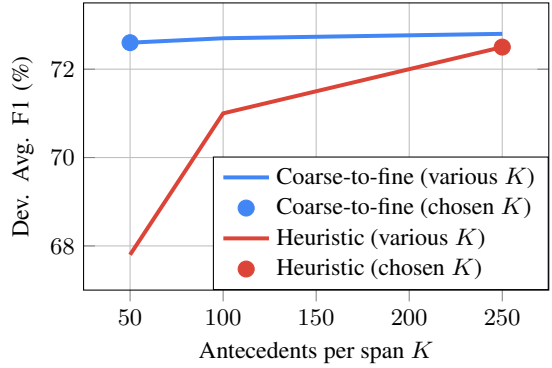

\subsection{Coarse-to-fine antecedent pruning}
We instead propose a coarse-to-fine approach that can be learned end-to-end and does not establish an a priori maximum coreference distance. The key component of this coarse-to-fine approach is an alternate bilinear scoring function:
\begin{align}
    \bscore{i}{j} &= \V{g}_i^\top \M{W}{c}\;\V{g}_j
\end{align}
where $\M{W}{c}$ is a learned weight matrix. In contrast to the concatenation-based $\ascore{i}{j}$, the bilinear $\bscore{i}{j}$ is far less accurate. A direct replacement of $\ascore{i}{j}$ with $\bscore{i}{j}$ results in a performance loss of over 3 F1 in our experiments. However, $\bscore{i}{j}$ is much more efficient to compute. Computing $\bscore{i}{j}$ only requires manipulating matrices of size $M \times |\V{g}|$ and $M \times M$.

\begin{table*}[ht!]
\setlength{\tabcolsep}{0.3em}
\def\arraystretch{0.95}
\centering
\begin{tabularx}{\linewidth}{X c*{14}{c}}
\toprule
& \multicolumn{4}{c}{MUC} &  \multicolumn{4}{c}{\bcubed} & \multicolumn{4}{c}{\ceaf} & \\ 
 & Prec.  & Rec.  & F1 & \; & Prec. & Rec. & F1 & \;  & Prec.  & Rec.  & F1 & \; &  \multicolumn{1}{c}{Avg. F1} \\
\midrule
\newcite{martschat:2015}  & 76.7 & 68.1 & 72.2 && 66.1 & 54.2 & 59.6 && 59.5 & 52.3 & 55.7 && 62.5\\
\newcite{clark:2015}      & 76.1 & 69.4 & 72.6 && 65.6 & 56.0 & 60.4 && 59.4 & 53.0 & 56.0 && 63.0\\
\newcite{wiseman:2015}    & 76.2 & 69.3 & 72.6 && 66.2 & 55.8 & 60.5 && 59.4 & 54.9 & 57.1 && 63.4\\
\newcite{wiseman:2016}    & 77.5 & 69.8 & 73.4 && 66.8 & 57.0 & 61.5 && 62.1 & 53.9 & 57.7 && 64.2\\
\newcite{clark:2016a}     & 79.9 & 69.3 & 74.2 && 71.0 & 56.5 & 63.0 && 63.8 & 54.3 & 58.7 && 65.3\\ 
\newcite{clark:2016b}     & 79.2 & 70.4 & 74.6 && 69.9 & 58.0 & 63.4 && 63.5 & 55.5 & 59.2 && 65.7\\ 
\cmidrule{1-14}
\newcite{e2e-coref}& 78.4 & 73.4 & 75.8 && 68.6 & 61.8 & 65.0 && 62.7 & 59.0 & 60.8 && 67.2\\
~~+ \texttt{ELMo}~\cite{elmo} & 80.1 & 77.2 & 78.6 && 69.8 & 66.5 & 68.1 && 66.4 & 62.9 & 64.6 && 70.4 \\
~~~~+ hyperparameter tuning &80.7 & 78.8 & 79.8 && 71.7 & 68.7 & 70.2 && 67.2 & 66.8 & 67.0 && 72.3\\
\cmidrule{1-14}
~~~~~~+ coarse-to-fine inference &80.4 & \textbf{79.9} & 80.1 && 71.0 & \textbf{70.0} & 70.5 && 67.5 & \textbf{67.2} & 67.3 && 72.6\\
~~~~~~~~+ second-order inference & \textbf{81.4} & 79.5 & \textbf{80.4} && \textbf{72.2} & 69.5 & \textbf{70.8} && \textbf{68.2} & 67.1 & \textbf{67.6} && \textbf{73.0}\\
\bottomrule
\end{tabularx}

\caption{Results on the test set on the English CoNLL-2012 shared task. The average F1 of \muc, \bcubed, and \ceaf is the main evaluation metric. We show only non-ensembled models for fair comparison.}
\label{tab:test_results}
\end{table*}

Therefore, we instead propose to use $\bscore{i}{j}$ to compute a rough sketch of \emph{likely} antecedents. This is accomplished by including it as an additional factor in the model:
\begin{align}
\hspace{-10pt}\cscore{i}{j} &= \mscore{i} + \mscore{j} + \bscore{i}{j} + \ascore{i}{j}\hspace{-10pt}
\end{align}
Similar to the baseline model, we leverage this additional factor to perform an additional beam pruning step. The final inference procedure involves a three-stage beam search:
\paragraph{First stage} Keep the top $M$ spans based on the mention score $\mscore{i}$ of each span.
\paragraph{Second stage} Keep the top $K$ antecedents of each remaining span $i$ based on the first three factors, $\mscore{i} + \mscore{j} + \bscore{i}{j}$.
\paragraph{Third stage} The overall coreference $\cscore{i}{j}$ is computed based on the remaining span pairs. The soft higher-order inference from Section~\ref{sec:higher_order} is computed in this final stage. 

While the maximum-likelihood objective is computed over only the span pairs from this final stage, this coarse-to-fine approach expands the set of coreference links that the model is capable of learning. It achieves better performance while using a much smaller $K$ (see Figure~\ref{fig:pruning}).

\section{Experimental Setup}
We use the English coreference resolution data from the CoNLL-2012 shared task~\cite{pradhan:2012} in our experiments. The code for replicating these results is publicly available.\footnote{\url{https://github.com/kentonl/e2e-coref}} 

Our models reuse the hyperparameters from ~\newcite{e2e-coref}, with a few exceptions mentioned below. In our results, we report two improvements that are orthogonal to our contributions.
\begin{itemize}
    \item We used embedding representations from a language model~\cite{elmo} at the input to the LSTMs (\texttt{ELMo} in the results).
    \item We changed several hyperparameters:
    \begin{enumerate}
        \item increasing the maximum span width from 10 to 30 words.
        \item using 3 highway LSTMs instead of 1.
        \item using GloVe word embeddings ~\cite{glove} with a window size of 2 for the head word embeddings and a window size of 10 for the LSTM inputs.
    \end{enumerate}
\end{itemize}
The baseline model considers up to 250 antecedents per span. As shown in Figure~\ref{fig:pruning}, the coarse-to-fine model is quite insensitive to more aggressive pruning. Therefore, our final model considers only 50 antecedents per span.

On the development set, the second-order model ($N=2$) outperforms the first-order model by 0.8 F1, but the third order model only provides an additional 0.1 F1 improvement. Therefore, we only compute test results for the second-order model.

\section{Results}
We report the precision, recall, and F1 of the the \muc, \bcubed, and \ceaf metrics using the official CoNLL-2012 evaluation scripts. The main evaluation is the average F1 of the three metrics.

Results on the test set are shown in Table~\ref{tab:test_results}. We include performance of systems proposed in the past 3 years for reference. The baseline relative to our contributions is the span-ranking model from ~\newcite{e2e-coref} augmented with both \texttt{ELMo} and hyperparameter tuning, which achieves 72.3 F1. Our full approach achieves 73.0 F1, setting a new state of the art for coreference resolution.

Compared to the heuristic pruning with up to 250 antecedents, our coarse-to-fine model only computes the expensive scores $\ascore{i}{j}$ for 50 antecedents. Despite using far less computation, it outperforms the baseline because the coarse scores $\bscore{i}{j}$ can be computed for all antecedents, enabling the model to potentially predict a coreference link between any two spans in the document. As a result, we observe a much higher recall when adopting the coarse-to-fine approach.

We also observe further improvement by including the second-order inference (Section~\ref{sec:higher_order}). The improvement is largely driven by the overall increase in precision, which is expected since the higher-order inference mainly serves to rule out inconsistent clusters. It is also consistent with findings from \newcite{martschat:2015} who report mainly improvements in precision when modeling latent trees to achieve a similar goal.

\section{Related Work}
In addition to the end-to-end span-ranking model~\cite{e2e-coref} that our proposed model builds upon, there is a large body of literature on coreference resolvers that fundamentally rely on scoring span pairs~\cite{ng:2002,bengtson:2008,denis2008specialized,fernandes:2012,durrett:2013,wiseman:2015,clark:2016b}.

Motivated by structural consistency issues discussed above, significant effort has also been devoted towards cluster-level modeling. Since global features are notoriously difficult to define~\cite{wiseman:2016}, they often depend heavily on existing pairwise features or architectures~\cite{bjorkelund:2014,clark:2015,clark:2016a}. We similarly use an existing pairwise span-ranking architecture as a building block for modeling more complex structures. In contrast to ~\newcite{wiseman:2016} who use highly expressive recurrent neural networks to model clusters, we show that the addition of a relatively lightweight gating mechanism is sufficient to effectively model higher-order structures.

\section{Conclusion}
We presented a state-of-the-art coreference resolution system that models higher order interactions between spans in predicted clusters. Additionally, our proposed coarse-to-fine approach alleviates the additional computational cost of higher-order inference, while maintaining the end-to-end learnability of the entire model.

\subsection*{Acknowledgements}
The research was supported in part by DARPA under the DEFT program (FA8750-13-2-0019), the ARO (W911NF-16-1-0121), the NSF (IIS-1252835, IIS-1562364), gifts from Google and Tencent, and an Allen Distinguished Investigator Award. We also thank the UW NLP group for helpful conversations and comments on the work.

\bibliography{main}

\begin{thebibliography}{}
\expandafter\ifx\csname natexlab\endcsname\relax\def\natexlab#1{#1}\fi

\bibitem[{Bengtson and Roth(2008)}]{bengtson:2008}
Eric Bengtson and Dan Roth. 2008.
\newblock Understanding the value of features for coreference resolution.
\newblock In {\em EMNLP\/}.

\bibitem[{Bj\"{o}rkelund and Kuhn(2014)}]{bjorkelund:2014}
Anders Bj\"{o}rkelund and Jonas Kuhn. 2014.
\newblock Learning structured perceptrons for coreference resolution with
  latent antecedents and non-local features.
\newblock In {\em ACL\/}.

\bibitem[{Clark and Manning(2015)}]{clark:2015}
Kevin Clark and Christopher~D. Manning. 2015.
\newblock Entity-centric coreference resolution with model stacking.
\newblock In {\em ACL\/}.

\bibitem[{Clark and Manning(2016{\natexlab{a}})}]{clark:2016b}
Kevin Clark and Christopher~D. Manning. 2016{\natexlab{a}}.
\newblock Deep reinforcement learning for mention-ranking coreference models.
\newblock In {\em EMNLP\/}.

\bibitem[{Clark and Manning(2016{\natexlab{b}})}]{clark:2016a}
Kevin Clark and Christopher~D. Manning. 2016{\natexlab{b}}.
\newblock Improving coreference resolution by learning entity-level distributed
  representations.
\newblock In {\em ACL\/}.

\bibitem[{Denis and Baldridge(2008)}]{denis2008specialized}
Pascal Denis and Jason Baldridge. 2008.
\newblock Specialized models and ranking for coreference resolution.
\newblock In {\em EMNLP\/}.

\bibitem[{Durrett and Klein(2013)}]{durrett:2013}
Greg Durrett and Dan Klein. 2013.
\newblock Easy victories and uphill battles in coreference resolution.
\newblock In {\em EMNLP\/}.

\bibitem[{Fernandes et~al.(2012)Fernandes, Dos~Santos, and
  Milidi{\'u}}]{fernandes:2012}
Eraldo~Rezende Fernandes, C{\'\i}cero~Nogueira Dos~Santos, and Ruy~Luiz
  Milidi{\'u}. 2012.
\newblock Latent structure perceptron with feature induction for unrestricted
  coreference resolution.
\newblock In {\em CoNLL\/}.

\bibitem[{Hochreiter and Schmidhuber(1997)}]{lstm}
Sepp Hochreiter and J{\"u}rgen Schmidhuber. 1997.
\newblock {Long Short-term Memory}.
\newblock {\em Neural computation\/} .

\bibitem[{Lee et~al.(2017)Lee, He, Lewis, and Zettlemoyer}]{e2e-coref}
Kenton Lee, Luheng He, Mike Lewis, and Luke~S. Zettlemoyer. 2017.
\newblock End-to-end neural coreference resolution.
\newblock In {\em EMNLP\/}.

\bibitem[{Martschat and Strube(2015)}]{martschat:2015}
Sebastian Martschat and Michael Strube. 2015.
\newblock Latent structures for coreference resolution.
\newblock {\em TACL\/} .

\bibitem[{Ng and Cardie(2002)}]{ng:2002}
Vincent Ng and Claire Cardie. 2002.
\newblock Identifying anaphoric and non-anaphoric noun phrases to improve
  coreference resolution.
\newblock {\em Computational linguistics\/} .

\bibitem[{Pennington et~al.(2014)Pennington, Socher, and Manning}]{glove}
Jeffrey Pennington, Richard Socher, and Christopher~D. Manning. 2014.
\newblock Glove: Global vectors for word representation.
\newblock In {\em EMNLP\/}.

\bibitem[{Peters et~al.(2018)Peters, Neumann, Iyyer, Gardner, Clark, Lee, and
  Zettlemoyer}]{elmo}
Matthew~E. Peters, Mark Neumann, Mohit Iyyer, Matt Gardner, Christopher Clark,
  Kenton Lee, and Luke Zettlemoyer. 2018.
\newblock Deep contextualized word representations.
\newblock In {\em HLT-NAACL\/}.

\bibitem[{Pradhan et~al.(2012)Pradhan, Moschitti, Xue, Uryupina, and
  Zhang}]{pradhan:2012}
Sameer Pradhan, Alessandro Moschitti, Nianwen Xue, Olga Uryupina, and Yuchen
  Zhang. 2012.
\newblock Conll-2012 shared task: Modeling multilingual unrestricted
  coreference in ontonotes.
\newblock In {\em CoNLL\/}.

\bibitem[{Wiseman et~al.(2016)Wiseman, Rush, and Shieber}]{wiseman:2016}
Sam Wiseman, Alexander~M Rush, and Stuart~M Shieber. 2016.
\newblock Learning global features for coreference resolution.
\newblock In {\em NAACL-HLT\/}.

\bibitem[{Wiseman et~al.(2015)Wiseman, Rush, Shieber, and
  Weston}]{wiseman:2015}
Sam Wiseman, Alexander~M. Rush, Stuart~M. Shieber, and Jason Weston. 2015.
\newblock Learning anaphoricity and antecedent ranking features for coreference
  resolution.
\newblock In {\em ACL\/}.

\end{thebibliography}
\bibliographystyle{acl_natbib}

\end{document}